\documentclass[acmsmall,screen]{acmart}

\AtBeginDocument{%
  }

\setcopyright{acmlicensed}
\copyrightyear{2025}
\acmYear{2025}
\acmDOI{XXXXXXX.XXXXXXX}
\acmConference[CSCW '26]{The ACM Conference on Computer-Supported Cooperative Work and Social Computing}{}{}
\acmISBN{978-1-4503-XXXX-X/2018/06}

\usepackage{color}
\usepackage{verbatim}

\newcommand{\newt}[1]{\textcolor{black}{#1}}

\begin{document}

\title{Examining the Prevalence and Dynamics of AI-Generated Media in Art Subreddits}

\author{Hana Matatov}
\affiliation{%
  \institution{Technion - Israel Institute of Technology}
  \country{Israel}
}
\author{Marianne Aubin Le Qu\'er\'e}
\affiliation{%
  \institution{Cornell Tech}
  \country{United States}
}
\author{Ofra Amir}
\affiliation{%
  \institution{Technion - Israel Institute of Technology}
  \country{Israel}
}
\author{Mor Naaman}
\affiliation{%
  \institution{Cornell Tech}
  \country{United States}
}









\begin{abstract}
Broadly accessible generative AI models like Dall-E have made it possible for anyone to create compelling visual art.
In online communities, the introduction of AI-generated content (AIGC) may impact social dynamics, for example causing changes in who is posting content, or shifting the norms or the discussions around the posted content if posts are suspected of being generated by AI.
We take steps towards examining the potential impact of AIGC on art-related communities on Reddit. 
We distinguish between communities that disallow AI content and those without such a direct policy.
We look at image-based posts in these communities where the author transparently shares that the image was created by AI, and at comments in these communities that suspect or accuse authors of using generative AI. 
We find that AI posts (and accusations) have played a surprisingly small part in these communities through the end of 2023, accounting for fewer than 0.5\% of the image-based posts. 
However, even as the absolute number of author-labeled AI posts dwindles over time, accusations of AI use remain more persistent. 
We show that AI content is more readily used by newcomers and may help increase participation if it aligns with community rules.
However, the tone of comments suspecting AI use by others has become more negative over time, especially in communities that do not have explicit rules about AI.
Overall, the results show the changing norms and interactions around AIGC in online communities designated for creativity.  
\end{abstract}

\maketitle

\section{Introduction}
Online communities are grappling with how to respond to the availability of AI-generated content (AIGC) online.
On the one hand, communities that forbid the use of AI often cite that content quality may be lower, creative perspectives will be homogenized, and community values could be compromised~\cite{lloyd2023there, frosio2023should, radivojevic2024human, lloyd2025airules}.
Conversely, communities that choose not to forbid AI may be calling upon arguments that AI can be used in ways that help democratize certain hobbies, spark creativity, and may increase participation~\cite{gero2023social, lloyd2023there, panchanadikar2024new}.
As online communities wrestle with these choices and strategies, characterizing the dynamics and behaviors of community members, especially in communities that differ in their rules toward AI, can inform future community actions. 
We explore how this tension in explicitly restricting AI plays out within art subreddit communities, which are highly creative, have specific community values, and where members may have broad access to AI for image generation.

AIGC has been described as a ``paradigm shift in content creation and knowledge representation''~\cite{wang2023survey}, and already has significant impact on online information.
Across different fields, emergent work has attempted to quantify the prevalence and responses to AIGC online, from impact to academic peer review and synthetic news articles, to online crowdwork ~\cite{liang2024monitoring,latona2024ai, veselovsky2023artificial, christoforou2024generative, hanley2024machine,vykopal2023disinformation}.
However, there is a need to understand how AIGC may impact online social ecosystems, since these are often based on community values and authenticity in ways that may be called into question by the widespread adoption of AIGC~\cite{lloyd2023there,lloyd2025airules}.
In this study, we tackle this question as it relates to online art communities.

We are interested in how generative AI changes some core aspects of online community life.
The first aspect is content production. 
Online communities lean on the active production of relevant content through posts and discussions to fulfill their purpose and maintain their perceived popularity~\cite{gero2023social,panchanadikar2024new,atcheson2024not}.
With generative AI, we may expect shifts in content production, both in terms of quantity and perceived quality, as AI tools change who can create and what they create~\cite{wei2024understanding,zhou24art}. 
\newt{Establishing the prevalence of AI-generated posts in online communities is crucial to understand the broader impact of these technologies in online social experiences.}
The second aspect is participation. Online communities rely on participation, and in this regard, practices, policies, and norms around participation barriers often dictate the dynamics and even success of communities~\cite{atcheson2024not,frosio2023should, radivojevic2024human, lloyd2025airules}. AIGC may influence who participates, potentially lowering barriers for newcomers or conversely, deterring some contributors~\cite{kraut2012challenges,wei2024understanding}.
The third aspect is the communication and trust between community members. In online communities, it is often critical that members trust each other and have a positive exchange~\cite{mazzone2019art}.
AIGC may alter how members engage with each other and the overall social dynamics in
the community~\cite{guo2024exploring,garcia2024paradox}. 
We therefore address the following research questions:
\begin{itemize}
\item RQ1: How prevalent are AIGC posts within online art communities, and how does the volume of such posts evolve over time?
\item RQ2: Does AIGC facilitate the participation of new creators in these communities, and do they remain active contributors over time?
\item RQ3: How do community members react to and engage with posts containing media suspected of being AI-generated?
\end{itemize}
Moreover, we examine these research questions in the context of the communities' approach towards generative AI. We examine whether these AI-related participation and community dynamics differ between art communities that forbid AIGC and those that remain neutral towards AI-generated content. 

To this end, we collect a list of 57 art-related subreddit communities, which we label as \textit{AI-neutral} or \textit{AI-disallowed} based on their stance towards AI posts.
Having all posts in these subreddits, spanning from the beginning of each subreddit through the end of 2023~\cite{torrent}, we retrieve and classify 28,405 image-based posts where the poster has either admitted to generating the image with AI (which we call a \textit{Transparent AI post}), or a commenter has accused the image of being AI-generated (\textit{Suspected AI post}).
We then analyze the number of AI-generated posts made to these communities over time, the trajectories of newcomers to these communities who post AI-generated images, and the speed and impact of AI-suspecting comments on posts.

Overall, we find that the prevalence of posts made to these art-based communities where the authors transparently self-report AI-generated use, or where commenters raise suspicion about the use of AI, is surprisingly low, affecting less than 0.5\% of image-based posts in our dataset.  
We find that the production of \textit{transparent} AI-based image posts started increasing in early 2022, peaked in August 2022 and has decreased since, while comments \textit{suspecting} the use of AI started rising only months later, and may not have declined as much. 
These trends were similar between communities that disallow AI and those that do not explicitly do so. 
Further, we identify that transparent AI posts are more likely to be made by newcomers to a community, and that newcomers who initially transparently share AI-generated images often continue to share AI-generated images, indicating that their interest in these communities has been enabled by the availability of these technologies.
Finally, we find that image-based posts that are suspected of using AI have been receiving exceedingly negative comments over the period of the data collection.

More broadly, our findings set the stage for understanding the prevalence, impact, and reception of AIGC in online visual art communities. 

\section{Background}
Our research focuses on online art communities, where we investigate the discussions about AI-generated images, specifically AI-generated art. 
The work therefore relates to extensive prior work in online communities, and emerging work on AI-generated content, particularly images.

\subsection{Online Communities}
Online communities have been an important object of study for internet researchers.
One theory that applies to online communities, and is especially relevant to the work here, is the concept of communities of practice (CoP)~\cite{wenger1999communities,wenger1996we,lave1991situating,mcdonald2016communities,wenger2002seven}, where groups of people who share a common interest come together to explore these ideas, share, and improve their practice.
CoP theory emphasizes that members engage in collective learning through shared practices, discussions, and activities, as is the case in Reddit art-based communities. 
Broadly speaking, online communities often wrestle with tradeoffs between different priorities, such as growth and attracting newcomers, sustainability, content quality, and fostering a sense of virtual belonging~\cite{kairam2022social, lin2017better, hwang2021people, kraut2012challenges}. 
Recent work on Reddit mapped subreddits into five archetype, which relate to different ways they produce a sense of community~\cite{prinster2024community}.
In this work, we are interested in art communities, which are probably most aligned with the \textit{Content Generation} archetype from that study, and aim to understand how such a community of practice might be impacted by AIGC. 

Indeed, the recent emergence of AIGC may present a boon for online communities, but could also put community values under threat.
The Diffusion of Innovations theory, formulated by Rogers~\cite{rogers2014diffusion,rogers1979diffusion}, focuses on the process by which individuals adopt new innovations and ideas, suggesting that different communities may also adapt innovations in diverse ways.
Weld et al.~\cite{weld2024making} identify that Reddit communities put emphasis on diverse values, and often prioritizing one comes at the cost of another -- for example, the quality of content can be in direct conflict with the need to grow communities and be self-sustaining.
Prior work has identified that the choice of whether to allow or disallow AI can force subreddit moderators to contend with which of these values to codify as rules~\cite{lloyd2023there,lloyd2025airules}.
These decisions can have consequences beyond individual subreddits; recent work has demonstrated that simple changes like exposing moderation decisions significantly impact the behaviors of users on the Reddit platform~\cite{jhaver2024bystanders}.
As communities content with the decision of how to moderate AI-generated content, this study provides initial evidence for how these decision may shape participation in online communities.

\subsection{AI-Generated Images}

Text-to-image models, such as DALL-E,
Stable Diffusion,
and Midjourney,
have exhibited exceptional performance in generating both abstract and photorealistic images. 
Prior to the advent of these models, Generative Adversarial Networks (GAN)-based models were employed to generate images~\cite{brock2018large,goodfellow2014generative,zhang2021cross}. 
With such popular image-generation tools available since 2022, the concept of AI-generated content have been increasingly impacting online communities. 

Since these tools are so advanced, one key challenge is the automatic detection or manual identification of AI-generated images. 
Efforts to determine whether an image was generated by generative models reveal the challenges humans face in accurately identifying AI-generated images~\cite{lu2024seeing,nightingale2022ai}, with a misclassification rate of 38.7\%~\cite{lu2024seeing}. Furthermore, even the top-performing AI-generated image detection model achieves a misclassification rate of 13\%~\cite{lu2024seeing}.

A few papers have specifically examined the detection of AI-generated art~\cite{ha2024organic,wang2023benchmarking}. Ha et al.~\cite{ha2024organic} showed that an approach combining human experts and automatic detectors was most effective, while non-artist crowdworkers generally were unable to discern between human-created art and AI-generated images.

Given this technological ability to produce images, including art, and  the inability of humans or machines to discern generated images, it is natural to expect that online communities might be impacted by AIGC. 

\subsection{AI-generated Art and Online Communities}
AI-generated art is not just as an attempt for fabrication or low-effort creation.
Several papers make the case for AI-generated art creations as authentic and creative~\cite{mazzone2019art,garcia2024paradox}.
Garcia~\cite{garcia2024paradox} specifically investigates AI-generated art produced by advanced technologies like DALL-E.
The paper provides a comprehensive analysis of the challenges associated with generative AI in art, including issues of authenticity, intellectual property, ethical concerns, the influence on traditional art practices, and market oversaturation. 

The discussions of challenges and opportunities in AI art extend to the context of online art communities.
Communities of practice may have reasons to adopt or reject AI-generated art. 
For example, AI art may be seen as not contributing to improving their practice, or lacking creativity. 
On the other hand, AI art could be viewed as a new avenue for contribution, or a way to lower the barriers for newcomers to participate~\cite{kraut2012challenges}.



Some recent works directly explore the impact of AI-generated images on different artistic communities ~\cite{wei2024understanding, guo2024exploring, zhou24art}.
Wei and Tyson~\cite{wei2024understanding} examined the spread of AI-generated images on Pixiv, a Japanese online community for artists, where artists are required to explicitly label their submissions as either human-generated or AI-generated. 
The paper reveals that the introduction of AI led to a 50\% increase in new artworks, but no corresponding increase in the number of views or comments. Additionally, the introduction of AI prompted an increase in the number of new creators, but a 4.3\% decrease in new creators of human-generated content. 
Another exploratory work~\cite{guo2024exploring} focused on the artist community DeviantArt which also explicitly allows AIGC. 
Despite DeviantArt's requirement for users to indicate the use of AI techniques, the authors identified AI-generated artworks that did not explicitly disclose AI usage, exposing the complicated motivation of contributors. 
Zhou and Lee~\cite{zhou24art} looked at data from an undisclosed online community 
focused on the creativity and productivity of artists who adopted generative AI, providing evidence for a broadening of participation and success in the community examined. 

The papers mentioned above examined stand-alone online communities dedicated to art, with centralized policies and rules. 
In this work, we focus on Reddit, and particularly on the multiple decentralized, mostly-self governed~\cite{fiesler2018reddit} communities on this site, and especially those dedicated to art.  
Recent work has shown that subreddit moderators are already grappling with AIGC~\cite{lloyd2023there,lloyd2025airules}.
Subreddit moderators are creating rules that regulate AI use in their communities, clarify norms, and enforce policies.
By leveraging the organizational structure of Reddit's subreddits, we can analyze data based on the distinct characteristics of different communities, including their own rules --- namely, if the community explicitly disallows AI, or is neutral towards it in its rules.
\newt{Our analysis also covers questions around how AIGC may impact submission volume, the type of participants that contribute to online communities, and the community response to AIGC content in relation to different community moderation strategies. }

\newt{Overall, while prior work emphasized moderation approaches and governance mechanisms~\cite{lloyd2023there,lloyd2025airules}, our work complements these approaches by examining how AIGC interacts with the experience and practices of users -- specifically how users disclose, perceive, and engage with such content.}

\section{Dataset}
To investigate online discussions within art communities regarding AI-generated images, we undertook a careful process to select data from Reddit communities (subreddits) dedicated to art.
Then, to collect posts and comments related to AI-generated images, we used an initial keyword-based filtering using a list of generative AI keywords, presented in this section, and LLM-based labeling, detailed in the next section. 

%

Since there is no known existing list of art-related subreddits from prior work, we used the Reddit API combined with several external sources to construct an initial comprehensive list.
These sources include art-related subreddits retrieved using the PRAW API's search function\footnote{\url{https://praw.readthedocs.io/en/latest/index.html}}, a Reddit page detailing Reddit’s largest communities\footnote{\url{https://www.reddit.com/best/communities/1/}}, and an unofficial article describing the best art subreddits\footnote{\url{https://www.reddit.com/r/redditlists/comments/141nga/list_of_art_subreddits/}}.
This initial step yielded a list of 276 subreddits. 
We further narrowed this initial list of subreddits by manually examining the description of each subreddit to ascertain its relevance to art. 
When descriptions were insufficient for a decision, we examined a random sample of five posts from the subreddit.
A subreddit was considered to be related to art if it featured visual art, regardless of whether the art posted in the subreddit is created by the authors of the posts, and could be focused on a specific genre of art, or be more general. 
Subreddits dedicated to memes, videos, or nudity were excluded. 
This elimination step resulted in 141 relevant subreddits.
Finally, since we wanted to focus on AI use and response to AI in broader art communities, we excluded
the ``aiArt'' subreddit, resulting in 140 relevant subreddits.

After identifying the relevant subreddits, we proceeded to collect corresponding posts.
We utilized publicly available data hosted on Academic Torrents, which contains files for the top 40,000 subreddits, spanning from the beginning of the creation of each subreddit through the end of 2023~\cite{torrent}.
This data is mostly building on the Pushshift API~\cite{baumgartner2020pushshift}. 
Each data record corresponds to the Pushshift structure\footnote{\url{https://github.com/pushshift/api}}, including fields such as ID, date, title, author, and attached media. 
Given that these torrents only cover the top 40,000 subreddits, we successfully obtained torrents for 57 subreddits from our list, resulting in a final list of 57 subreddits for our analysis (see Appendix~\ref{list_subreddits}).

With the post and comment data from the subreddit submissions, we set out to identify potential discussions about generative AI by filtering for submissions that include media items, and by using a list of AI-related keywords.
First, we focused only on posts that included a media item, such as an image or video.
This focus allows us to analyze discussions related to specific AI-generated art instances, rather than general discussions about art and generative AI. 
Second, we filtered these posts using a list of keywords likely to appear in discussions about AI-generated images. 
We began with an initial list of keywords and searched for them within our set of subreddits. 
This initial list included various keywords, starting from general phrases such as ``generative art'' and ending with specific text-to-image tools names.
By examining the discussions in the retrieved posts, we iteratively refined, filtered, and expanded the list of keywords.
Each of these keywords might be a word or a phrase, and mostly includes combinations with the phrases ``AI'' or ``artificial intelligence'', or names of text-to-image tools for creating AI-generated images. 
This keyword selection process was done iteratively and purposefully to achieve a robust set, resulting in a final set of 68 keywords (see Appendix~\ref{list_keywords}).
We then identified occurrences of these 68 keywords in the titles, bodies, and comments of the posts in our 57 subreddits. 
This search yielded 28,864 posts or comments containing at least one of the keywords.

We excluded from our dataset posts or comments that were obviously published by bots or spammers. To this end, we identified all comments that were repeated more than five times across all subreddits, and searched for authors with usernames containing the term ``bot''. We categorized these authors into three groups: information bots (such as ``WikiSummarizerBot'' and ``RemindMeBot''), moderation bots (such as ``AutoModerator'' and ``art\_moderator\_bot''), and spammers (mostly users attempting to sell their services). 
We purposefully \textit{kept} five-times-or-more repeated comments that were posted by subreddit moderators. 
We retained these data entities since the moderators' comments indicate the post may include a discussion we are interested in, i.e. a discussion about AI-generated images.

Our final dataset includes 28,405 text items related to AI-generated images, from 14,574 distinct posts.
Of these, 6,393 (43.9\%) posts feature the keywords in the title or body, while 8,181 (56.1\%) posts have keywords only in the comments.  
The dataset includes several fields: ID, post title, post body, complete comments tree, author, timestamp, detected keywords, subreddit name, and more.
In the next sections, we enhance the dataset with the classification of the texts based on their relatedness to claims about AI-generative images, as well as the categorization of the included subreddits.
The dataset, with the classification labels and subreddit categories, will be made available upon publication of the paper.

\section{Classification of the Dataset Using LLMs}
Having gathered a dataset of potentially AI-relevant image-based posts based on keyword match, we further classified the posts using large language models (LLMs)~\cite{abburi2023generative,zhang2024pushing,pangakis2024knowledge}.
\newt{The overall goal of this classification task is to identify (a) \textbf{Transparent AI posts}, where the creator implies the image is AI-generated (b) \textbf{Suspected AI posts}, where a commenter implies the image is AI-generated. We approach this task from two aspects.}
First, since keyword matching can be imprecise, we use classifiers to filter the dataset to only retain posts and comments genuinely pertaining to AI-generated images.
Second, we differentiate between text that identifies the image in the post as being AI-generated, versus those that broadly mention AI-generated images.
\newt{Importantly, we do not attempt to establish whether the image is \textit{actually} AI-generated, since this task is beyond the scope of this paper which focuses on disclosures of and responses to AI-alleged images. While a perfect classifier for AI-generated images and art (a famously difficult challenge~\cite{ghiuruau2024distinguishing,epstein2023online}) would have been useful to substantiate these author and commenter claims about AI, the existence of such classifier would have negated the need to study disclosure and accusations altogether.}

\newt{Prior work has demonstrated the suitability of LLMs for text classification, when validated by human annotators~\cite{choksi2024under, dunivin2025scaling}.}
We thus developed textual prompts and utilized the ChatGPT API\footnote{\url{https://platform.openai.com/docs/api-reference/introduction}} (\texttt{GPT-4o} model).
Three distinct prompts were developed to correspond with each of the three types of text in our dataset: post's title or body, comments made by the author of the post, and comments by other users (see Appendix~\ref{llm_prompts_classification}).
These prompts were designed following the principles of chain-of-thought prompting~\cite{wei2022chain}.
Each prompt required the language model to provide reasoning before giving a classification decision.
Additionally, each prompt included nine examples from the dataset, implementing an in-context few-shot learning approach~\cite{wei2022chain}.

\newt{We used the LLMs to label each text according to one of three categories}: \textit{(A) Implies AI involvement}, when the text explicitly or implicitly suggests that AI tools were involved in generating the attached image; \textit{(B) Does not imply AI}, when the text mentions AI-generated images, but does not imply that the attached image is AI-generated (including texts that suggests that the attached image is not AI-generated); or \textit{(C) Unrelated to AI}, only when the text is completely unrelated to AI-generated images (regardless of whether the text refers to the image in the post)---in most cases, a false positive of the keyword match.
The prompts, along with examples of input texts and their corresponding classifications, are provided in Appendix~\ref{llm_prompts_classification}.

\newt{Following prior work~\cite{choksi2024under}, we manually evaluated the accuracy of each of the three prompts using three validation datasets.}
We randomly sampled 100 texts from each of the three types of text, and manually labeled \newt{the resulting 300 texts} according to the established coding.
Next, we classified the three validation datasets using the corresponding prompts.
\newt{The classifiers} achieved an accuracy of 93\% for the title of the posts, 82\% for comments made by the author of the post, and 84\% for comments from other accounts. 
In comparison, the baseline accuracy for the majority class \newt{(that is, the most frequent label among the three categories (A–C) within each validation set)} is 87\% for post titles, 49\% for comments made by the post’s author, and 55\% for comments from other accounts. 
Note that the majority class differs across the three validation sets: for post titles and comments made by the post’s author, the majority label is \textit{Implies AI involvement}, while for comments from other accounts, the majority label is \textit{Does not imply AI}.

\newt{Following validation, we classified} all data entities in the dataset using these prompts, following a process inspired by the QuaLLM framework~\cite{rao2024quallm}. 
According to this process we submitted the prompt, consisting in each case of the relevant class explanation 
and the in-context examples, with every 100 input texts at a time\footnote{Note that during the \newt{validation} phase we also compared the accuracy values obtained by \newt{varying the batch size (i.e., 10 instead of 100 texts at a time)}  and achieved identical results.}.
Similarly, we requested that the LLM return the final output as a dictionary, including both the reasoning and classifications for all these 100 texts.

As we expected, most of the keyword-matched texts in the dataset were indeed related to AI.
Amongst 6,393 posts, 5,144 were classified as implying the content was AI-generated in the title or body text. 
Otherwise, 792~posts were classified as mentioning AI-generated images, but without implying that the image in the post is AI-generated.
Amongst the 2,899 comments made by the author of the post that exist in our keyword-match dataset, 1,243 comments were classified as implying AI, and 1,050 as only mentioning AI.
Finally, for the collected comments by other users, amongst 19,113 keyword-matched comments, 7,322 were classified as implying AI was used in image creation, while 10,635 as only mentioning AI.

In this paper, we focus mainly on texts, either in posts or comments, that imply that the image attached to the post was AI-generated (the first category of labels above).
First, looking at people who make image posts in these communities, we identify on posts where the author implied or admitted that the image was AI-generated, whether in the title, or in one of the comments. 
We refer to posts where the creator itself implied the image was AI generated as \textbf{Transparent AI posts}.
There are 5,998 distinct posts like these where the creator provides this information in the title (5,144 times) and/or comments (1,243).
Second, we are interested in posts where other users or accounts make comments that refer in some way to the posted image as AI-generated.   
We refer to these posts where other accounts highlighted AI use as \textbf{Suspected AI posts}.
There were a total of 7,322 comments from other accounts claiming AI was used, made on 4,375 distinct \textit{Suspected AI posts}.
The overlap between the two sets of posts was small---576 posts where the author transparently posted AI and commenters specifically referred to AI generation in their comments.

\section{Categorization of the Communities}
We categorized the 57 subreddits into two types of communities: those with rules that explicitly ban AI images, and those whose rules do not mention AI.
Previous papers have explored the reactions of community moderators to AI-generated content, including focusing on the decisions of Reddit moderators to enact rules restricting the use of AI, as well as the responses of their communities~\cite{lloyd2023there,shankar2024advances}.
We set out to distinguish the different communities based on their expressed approach to AI content. 
Using the data provided via the PRAW API\footnote{\url{https://praw.readthedocs.io/en/latest/index.html}}, we examined each subreddit's API description, public description, and rules.
If any of these fields indicated that AI-generated content (or particularly AI-generated images or art) is not permitted, we categorized the subreddit as \textbf{AI-disallowed community}.
Otherwise, if the publication of AI-generated content was not mentioned in the subreddit's description or rules, it was categorized as \textbf{AI-neutral community}\footnote{In our dataset, no subreddit explicitly allows AI art in their description or rules.}.
Among the 57 subreddits, 24 explicitly forbid AI, while the remaining 33 subreddits do not reference AI in their descriptions and rules.

To obtain a more comprehensive understanding and baseline for the subreddits in each category, we utilized torrents~\cite{torrent} to count the total number of posts that included a media item in these communities, regardless of whether they contained the AI-related keywords.
We refer to these posts that included a media item, such as an image or video, as \textit{all image-based posts}.

\begin{table}[]
\centering
\begin{tabular}{|c|c|c|}
\hline
\textbf{} & \textbf{\begin{tabular}[c]{@{}c@{}}AI-disallowed \\ communities\end{tabular}} & \textbf{\begin{tabular}[c]{@{}c@{}}AI-neutral \\ communities\end{tabular}} \\ \hline
\textbf{Number of subreddits} & 24 & 33 \\ \hline
\textbf{\begin{tabular}[c]{@{}c@{}}Number of posts\end{tabular}} & 
\begin{tabular}[c]{@{}c@{}}4,920,664\end{tabular} & 
\begin{tabular}[c]{@{}c@{}}2,828,132\end{tabular} \\ \hline
\textbf{\begin{tabular}[c]{@{}c@{}}Number of all image-based posts\end{tabular}} & 
\begin{tabular}[c]{@{}c@{}}3,710,420\end{tabular} & 
\begin{tabular}[c]{@{}c@{}}2,183,990\end{tabular} \\ \hline
\textbf{\begin{tabular}[c]{@{}c@{}}Number of Transparent AI posts\end{tabular}} & 
\begin{tabular}[c]{@{}c@{}}4,458\end{tabular} & 
\begin{tabular}[c]{@{}c@{}}1,540\end{tabular} \\ \hline
\textbf{\begin{tabular}[c]{@{}c@{}}Number of Suspected AI posts\end{tabular}} & 
\begin{tabular}[c]{@{}c@{}}3,334\end{tabular} & 
\begin{tabular}[c]{@{}c@{}}1,041\end{tabular} \\ \hline
\textbf{\begin{tabular}[c]{@{}c@{}}Number of Suspected AI posts,\\ without former admission by the author\end{tabular}} & 
\begin{tabular}[c]{@{}c@{}}2,894\end{tabular} & 
\begin{tabular}[c]{@{}c@{}}905\end{tabular} \\ \hline
\end{tabular}
\caption{Characteristics of the \textit{AI-disallowed
} and \textit{AI-neutral} communities.
}
\label{table:communities_categories_table}
\end{table}

Table~\ref{table:communities_categories_table} presents a comparison of the volume and prevalence of various categories among the two community categories: those forbidding AI publications (\textit{AI-disallowed communities}), and those not referencing AI in their description and rules (\textit{AI-neutral communities}).
Although fewer communities explicitly forbid AI content (24 versus 33), these subreddits exhibit more activity, with approximately 4.9 million posts, and 3.7 million of them being image-based posts (posts including a media item), compared to 2.8 million posts and 2.2 million image-based posts in the \textit{AI-neutral communities}.
The table also suggests that subreddits that disallow AI have significantly more AI-related posts, with 4,458 posts where the author admitted that the image was AI-generated, and 3,334 posts where a commenter implied that the image was AI-generated, verses 1,540 and 1,041 respectively in the \textit{AI-neutral} subreddits.

This quantitative comparison reveals several unexpected findings.
Although the majority of posts of any kind within these subreddits are image-based, comprising 75\% in \textit{AI-disallowed} communities and 77\% in \textit{AI-neutral} communities, only a minimal percentage of the posts included texts implying AI involvement in creating the image.
Specifically, AI-claiming texts were presented only in 0.21\% of image-based posts in \textit{AI-disallowed} communities, and 0.12\% in \textit{AI-neutral} communities, including texts made by the author of the post (\textit{Transparent AI posts}) and those made by commenters (\textit{Suspected AI posts}).
Since AIGC was not prevalent before 2022, we recalculated the metric using data from January 1, 2022, onward. Across 2022–2023, \textit{Transparent} or \textit{Suspected AI posts} accounted for only 0.49\% of all image-based posts in both \textit{AI-disallowed} and \textit{AI-neutral communities}.
Furthermore, when AI involvement was mentioned in relation to the creation of the image, it was mostly by the content creators themselves, accounting for 57\% (4,458 out of 7,792) of cases in \textit{AI-disallowed communities}, and 60\% (1,540 out of 2,581) in \textit{AI-neutral communities}.
Finally, for 87\% (2,894 out of 3,334) of the \textit{Suspected AI posts} in the \textit{AI-disallowed communities}, i.e. posts where a non-author commenter implied AI involvement, the comment served as the first indication that the attached image was AI-generated, without the author of the post disclosing the use of AI.
The same percentage was observed in the second category as well. 
We explore these differences further in our analysis.

\section{Analyses and Findings}

In this section, we present a comprehensive analysis to address our research questions. Using the dataset we collected and refined in previous sections, we examine variations in volume, participation, and feedback patterns across \textit{AI-disallowed communities} and \textit{AI-neutral communities}.
This analysis highlights the distinctions between these two types of communities, as well as the differences in trends for the different types of posts --- \textit{Transparent AI posts} and \textit{Suspected AI posts}.

\subsection{Rise of AI Posts and Comments}

\begin{figure}[]
\includegraphics[height=0.9\textheight, width=1\columnwidth]{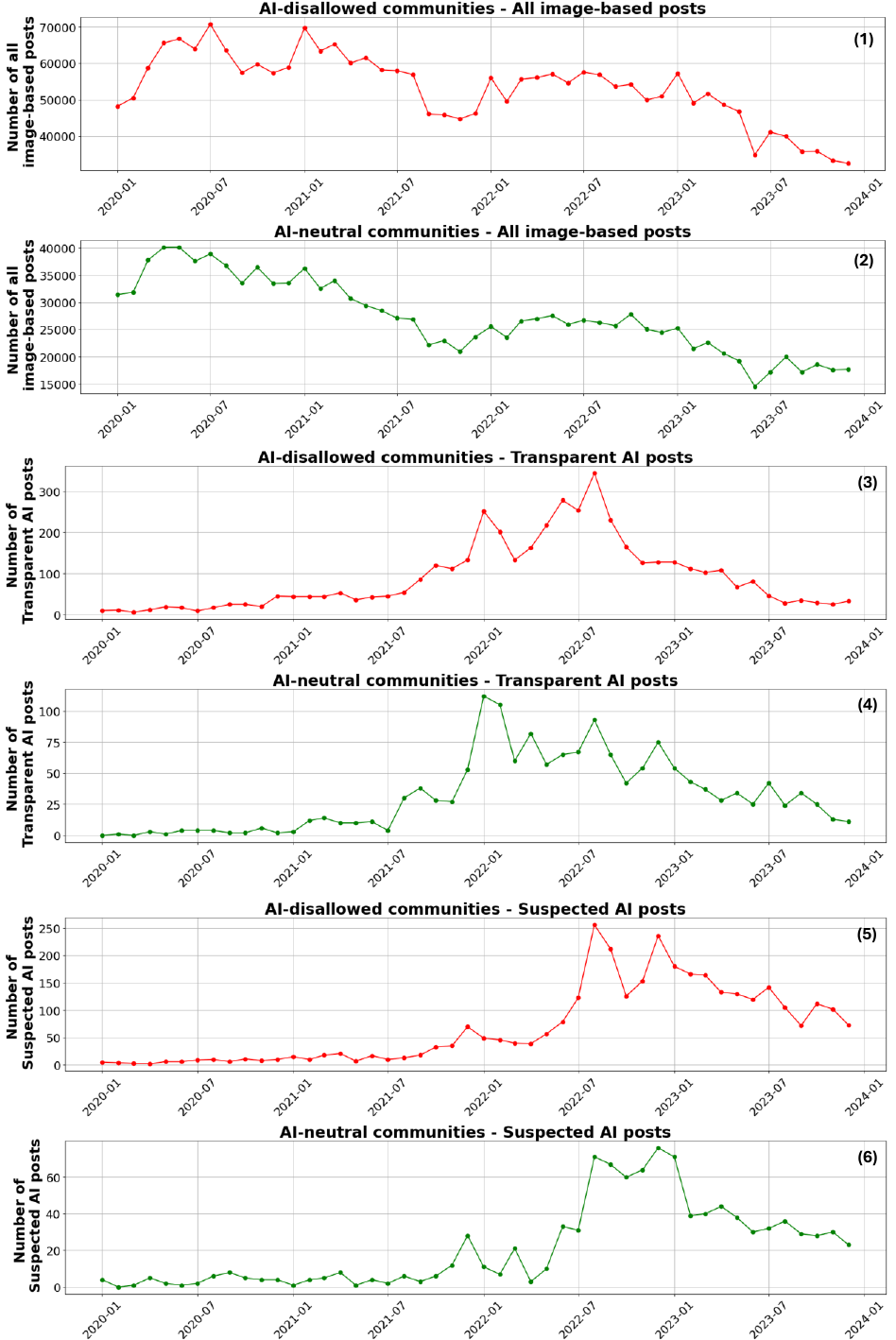}
\caption{The volume of posts over time in \textit{AI-disallowed} communities (red) and \textit{AI-neutral} communities (green).
The x-axis of the figure represents time (month), and the y-axis represents the number of posts. 
Plots 1-2 (top two) show the total volume of \textit{all image-based posts}
; plots 3-4 show \textit{Transparent AI posts}, and
plots (5-6) show the number of \textit{Suspected AI posts}.
}
\label{fig:timelines}
\end{figure}

To address our first research question, we conducted an analysis of post volume over time across different communities to understand how posting patterns shifted in response to the introduction and increasingly mainstream adoption of generative AI throughout 2022. 
Specifically, we examined temporal patterns of posts containing AI-generated images according to our classification, and compared these trends to the overall volume of image-based posts within the subreddits we examined.
We explore how these trends vary over time between the \textit{AI-disallowed communities} and the \textit{AI-neutral communities}. 
Figure~\ref{fig:timelines} shows the number of posts over time between both types of communities: \textit{AI-disallowed communities} in red, and \textit{AI-neutral communities} in green.
For brevity, the figure includes only posts published from the beginning of 2020 onward, as generative AI was not dominant prior to that period.
The figure presents the number of posts for various data segments, enumerating from the top panel: (1-2)
all \textit{image-based} posts, (3-4) \textit{Transparent AI posts} when authors admitted to posting AI-generated images, and (5-6) \textit{Suspected AI posts} when commenters implied that the image in the post is AI-generated.
For instance, Panel 2 shows that between March and August 2020, approximately 35,000 to 40,000 image-based posts were published in \textit{AI-neutral} communities (which do not mention AI posting in their community rules). 
In contrast, two years later (March-August 2022), only approximately 25,000 to 30,000 image-based posts were published in the same subreddits.  

Although the total number of posts differed between the two community categories, the temporal patterns exhibited similar trends. 
For example, panels~3 and~4 show that from January 2022 to September 2022, there was an increase in the number of \textit{Transparent AI posts}, peaking from early 2022 to August 2022.
Similarly, panels~5 and~6 show that in both types of communities, August 2022 to January 2023 was a peak of activity for comments about \textit{Suspected AI posts}.
The observed peak in August 2022 of posts with identified and suspected AI images aligns with significant developments in AI-generated imagery. 
OpenAI's DALL-E 2 was released in April 2022 and gained popularity with time, together with the launch of MidJourney's open beta version in July 2022, and the introduction of Stability AI's open-source model, Stable Diffusion, in August 2022.

While the overall number of image-based posts has been declining since August 2021  in both community types (panels 1-2), the number of \textit{Transparent AI posts} began to increase during the same period (panels 3-4).
Though analyzing general trends in image-based posts is beyond the scope of this work, this difference suggests that the rise in generative AI posts is not due to broader trends in posting volume across subreddits, but rather specific trends related to AI-generated images.

Figure~\ref{fig:timelines} exposes interesting differences between adoption of \textit{Transparent AI posts} in both community types (panels~3 and~4), and the response to posts in these communities measured via the \textit{Suspected AI posts} (panels~5 and~6).
In both communities, the number of \textit{Transparent AI posts} rose in early 2022 and reached a peak around August 2022, declining after that.
In contrast, \textit{Suspected AI posts} rose significantly around August 2022, and remained at relatively high levels longer after that.
This pattern suggests an initial phase where AI art creators increasingly shared their work, followed by a period during which participants in these subreddits became more aware and suspicious of AI-generated content, leading to sustained high levels of such claims by commenters.
We verified that the observed trends in the timelines were not disproportionately influenced by large subreddits.

To check this, we calculated the timelines using the percentage values for each subreddit (instead of volume values) -- an unbiased metric in terms of the varying volumes of subreddits -- averaged for each community category.
We revealed that the same trends persisted, indicating that the observed patterns were consistent and not dominated by the largest subreddits.
Furthermore, we examined the 12 subreddits with the highest number of posts, and observed similar trends in the timelines, despite the differences in post volumes.

A qualitative analysis showed that some of the earlier rise (Jan-May 2022) in \textit{Transparent AI posts} activity, in both types of communities, was due to one user. 
Notably, a single account 
was responsible for submitting 602 posts classified under ``Imply AI'' between mid-January 2022 and early May 2022.
Even without this user's data, our temporal comparison above and the differences between \textit{Transparent AI posts} and \textit{Suspected AI posts} remain. 

\subsection{Participation and Newcomers}
The previous sections focused on the volume of posts and discussions over time and in relation to other image posts in the same communities. 
In this section, we address our second research question
by examining \textit{participation} patterns in the creation of AI-related posts across the different art communities. 
\newt{In particular, we focus on newcomers -- users who make their first-ever image-sharing post to a certain subreddit.}
Is there evidence that AI-generated images make it easier for newcomers to contribute content -- and if so, do these newcomers stay and contribute over time?
Does this pattern diverge for those who are making \textit{Transparent AI posts} to \textit{AI-disallowed} versus \textit{AI-neutral} communities?

Overall, the participation of newcomers was significantly more emphasized in making \textit{Transparent AI posts} than the overall presence of newcomers in posting images to these communities.
In \textit{AI-disallowed communities}, newcomers usually publish 6.71\% of all image-based posts. This ratio is similar in \textit{AI-neutral communities}, with 6.51\% of newcomers publishing image-based posts. 
However, newcomers made as much as 25.15\% of \textit{Transparent AI posts} in \textit{AI-disallowed} communities during our period of analysis.
In \textit{AI-neutral communities}, newcomers made a similar 22.99\% of the \textit{Transparent AI posts}.
\newt{Chi-square tests showed that the proportion of posts made by newcomers is significantly different between \textit{Transparent AI posts} and image-based posts (\textit{AI-disallowed communities}: $\chi^2(1, N = 3,964,964) = 1796.71, p < 0.00001$; \textit{AI-neutral communities}: $\chi^2(1, N = 2,328,041) = 521.10, p < 0.00001$). 
}
This over-representation in newcomers posting AI-generated content suggests that this kind of content may encourage and broaden community participation.


We saw, then, that newcomers are participating more in creating \textit{Transparent AI posts}. But do newcomers who post AI content continue to contribute to the community? 
Here, we compared newcomers who posted AI content to all other newcomers to the community. 
We define \textit{Transparent AI newcomers} as those authors whose initial post in the subreddit was a \textit{Transparent AI post} (i.e. the author admitted the post includes an AI-generated image), and compared them to \textit{Image-based newcomers}, newcomers whose first contribution in the subreddit was any type of image-based post.
Subsequently, we calculated the percentage of newcomers who posted content from 2022 to 2023, and who posted again in the same subreddit at least one month after their initial post.

Overall, we found a significant drop off for AI-posting newcomers in \textit{AI-disallowed communities} compared to AI posters in \textit{AI-neutral communities} as well as compared to newcomers posting non-AI content. 
Of \textit{Transparent AI newcomers} in the \textit{AI-disallowed communities}, only 8.6\% remained active creators after a month. 
In the \textit{AI-neutral communities}, this rate of return was as high as 17.8\% for \textit{Transparent AI newcomers}. This observed difference also suggests that our categorization of subreddits accurately reflects their tolerance towards generative AI.
The rate of \textit{Transparent AI newcomers} remaining active for the \textit{AI-neutral communities} was similar for the overall rate for all new contributors (18.7\%) in these communities. 
This rate of newcomers who continued to participate was even higher, 23.5\%, for these general newcomers in the \textit{AI-disallowed communities} -- making the clear discouraging effect of the \textit{AI-disallowed communities} on newcomers posting AI even more apparent. 
The differences between  \textit{Transparent AI newcomers} and \textit{Image-based newcomers} were highly significant (Chi-square test, \( p < 0.00001 \)) for the \textit{AI-disallowed communities}, but not statistically significant for \textit{AI-neutral} communities. 

We did a similar analysis looking at newcomers who are still participating after three or six month. 
The patterns and differences between the communities remained similar, though, of course, with lower rates of continued participation as more time passed across all categories.
Overall, the findings suggest that in communities where the posting of generative AI content is not explicitly prohibited, newcomers who use AI as an entry point for participation in the subreddit continue to naturally participate over time.


The analysis above shows the often-low return rate of posters of AI-generated content, but we were also interested in whether the \textit{returning} newcomers continue to transparently post AI-generated content. 
In \textit{AI-disallowed communities}, when \textit{Transparent AI newcomers} returned after more than one month, 49.8\% of their subsequent image-based posts were also \textit{Transparent AI posts}.
This number was similar in \textit{AI-neutral communities} at 50.9\%, though, as we have seen above, there were relatively more returning posters in this group. 
Note that all these numbers build on counting \textit{Transparent AI posts} and thus are likely representing lower bounds for the proportion of \textit{actual} AI-generated content in the initial and subsequent posts, since
it is possible that 
AI-generated images are published without explicit acknowledgment by the poster.

Overall, the participation trends show that newcomers more heavily rely on posting AI content to participate in these communities, but that (as expected) their participation does not last long in \textit{AI-disallowed communities} \newt{(see section~\ref{sec:disc} for additional discussion).}
These trends may relate to the type of interactions around AI-generated content, which we look at next.

\subsection{Feedback, Interactions and Sentiment}
To address our third research question, we investigate the interactions---responses and comments---with identified or suspected AI-generated content, and how they are different in tone or dynamics from other content or across community types. 

First, we look at engagement levels with \textit{Transparent AI posts} across different communities.
We examined the engagement with image-based posts using the \textit{upvote ratio} as a metric, which represents the proportion of upvotes relative to the total number of votes (both upvotes and downvotes) received by a post.
For this analysis, we compared the distributions of upvote ratios in two distinct periods: (1) January-March 2022 and (2) January-March 2023.
These periods were selected to capture the differences over the span of a year, particularly given the release of significant text-to-image generation tools and the peak in activity observed towards the end of 2022, as discussed in previous sections.

\begin{figure}[]
\includegraphics[width=1\columnwidth]{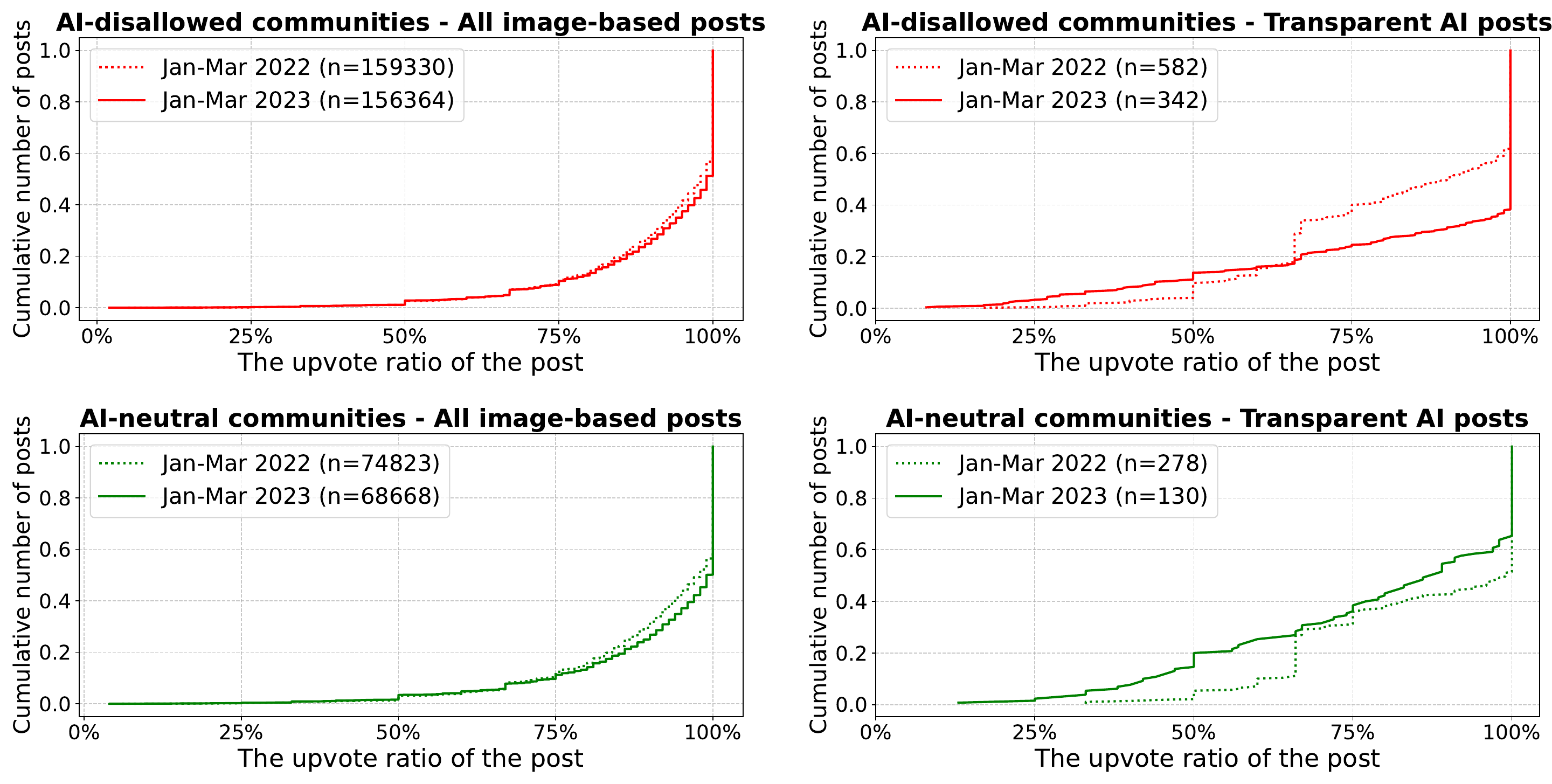}
\caption{The cumulative distribution of posts across different upvote ratio values (x-axis), and the corresponding cumulative percentage of posts (y-axis) for different kinds of post (left vs right) and communities (top vs bottom). 
}
\label{fig:upvote_ratio}
\end{figure}

Figure~\ref{fig:upvote_ratio} illustrates the cumulative distribution of posts across different upvote ratio ranges (X-axes), and the corresponding cumulative number of posts (Y-axes).
The left column panels display the distributions for all image-based posts in the analyzed subreddits, while the right panel focuses on \textit{Transparent AI posts}.
The upper row panels show posts within \textit{AI-disallowed} communities, whereas the lower panels present posts from \textit{AI-neutral} communities. 
The figure shows similar baseline response rates: the upvote ratio distribution for \textit{all image-based posts} across both community types was similar (top left and bottom left panels), and there were only minor differences (dotted vs. solid line) between 2022 and 2023 upvote ratios in both.

While the responses to all image posts remained stable, the responses to AI posts (panels on the right side of the figure) had shifted between 2022 and 2023 in both types of communities, but in different directions.
In \textit{AI-disallowed communities}, responses to \textit{Transparent AI Posts} were more positive in terms of upvote ratio in 2023 than in 2022. 
However, it is possible that this higher positivity is a result of these posts getting less total engagement in these communities overall.
On the other hand, in \textit{AI-neutral} communities, responses to \textit{Transparent AI Posts} in 2023 had lower upvote ratio than in 2022. 

We conducted a similar analysis for \textit{Suspected AI posts}, which refer to posts where non-author commenters claimed or implied that the image was generated using AI tools. 
The engagement with \textit{Suspected AI Posts} is entirely different than \textit{all image-based posts} or \textit{Transparent AI posts}, and cannot be directly compared to either. 
These posts generally have more engagement (comments and upvotes) than the other categories.
This difference has a couple of explanations. 
First, \textit{Suspected AI Posts} by definition have at least one comment, which already suggests different distributions. 
Second, the higher comparable engagement could be due to the fact that as posts become highly favored by the community's participants, it is increasingly likely that at least one commenter will attribute the image's creation to the use of generative AI tools.

\begin{figure}[]
\includegraphics[width=1\columnwidth]{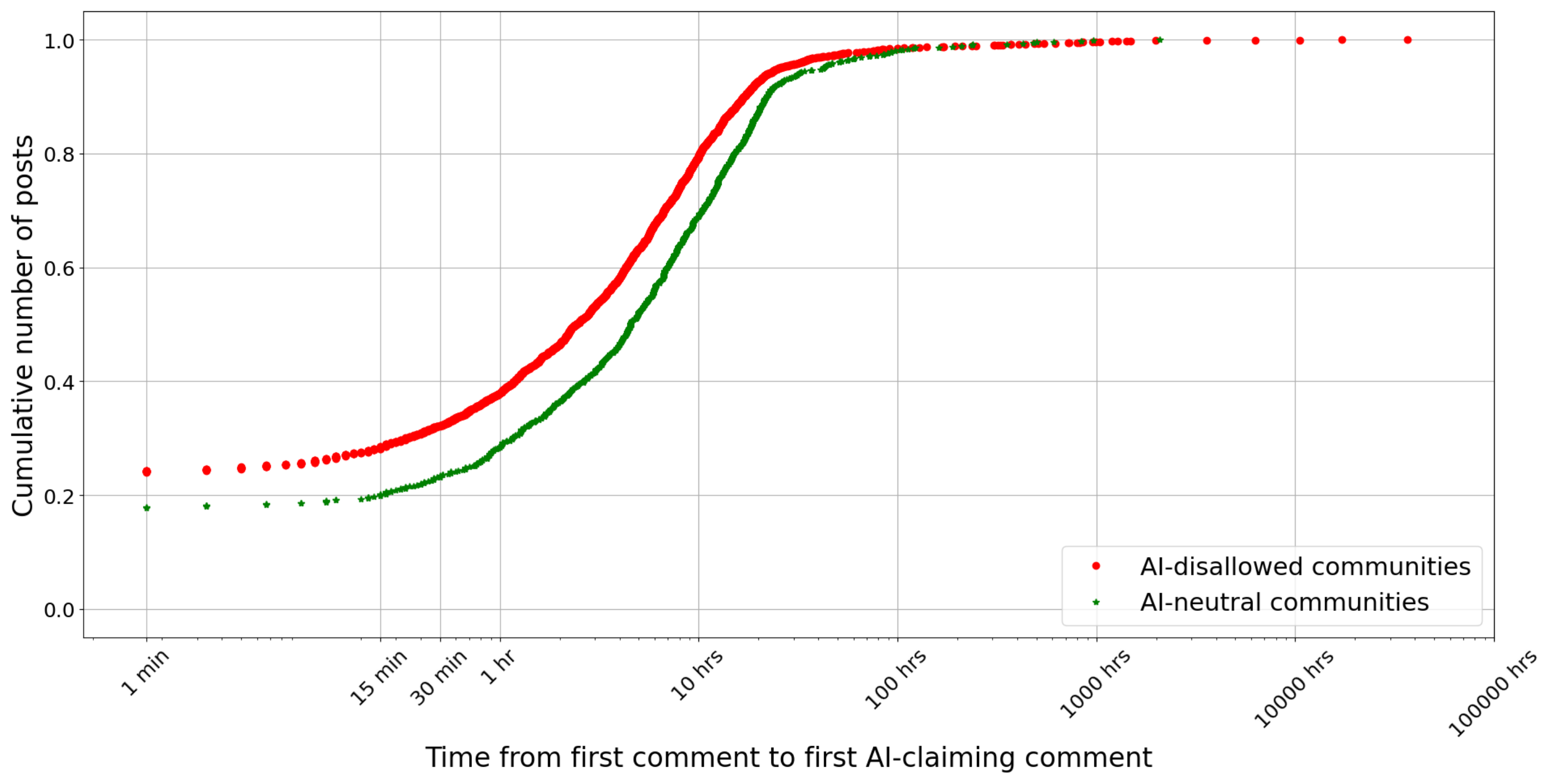}
\caption{The cumulative number of posts (y-axis) for time spanned between the publication of the first comment on a Suspected AI post and its first AI-claiming comment (x-axis).}
\label{fig:time_between_comments}
\end{figure}

Given this observation, we further focused on the dynamics around \textit{Suspected AI posts}, posts where a non-author commenter claimed or implied that the image attached to the post was generated using AI.
To better understand how do community members react to posts containing media suspected of being AI-generated, we looked at the speed with which these claims emerged and the content of these claims.
We began by calculating the time elapsed from the publication of the post until its first AI-claiming comment.
Our analysis revealed that approximately 20\% of the \textit{Suspected AI posts} had received their first AI-claiming comment within one hour from posting.
However, to compare between posts in the different types of communities, we had to control for the popularity of the post or its likelihood to be seen. 
To do that, we used the time window between the first comment on each \textit{Suspected AI post}, and the first AI-claiming comment on that post.
This \textit{between-comments} data is presented in Figure~\ref{fig:time_between_comments}, which shows the time between the comments (X-axis), and the cumulative number of posts (Y-axis).
For instance, the leftmost data point in the figure indicates that for both community types, approximately 20\% of the AI-claiming comments were made as the first comments on the posts (the X-axis tick mark of ``1~minute'' denoting ``within one minute'', and mostly captured the first comment). 
The figure shows that these comments appear more quickly in \textit{AI-disallowed communities}.
For example, around 40\% of the AI-claiming comments in these communities (red line) were published within one hour of the first comment. 
In \textit{AI-neutral communities}, this speed of action occurred for less than 30\% of the posts.
The figure also shows that after approximately 36 hours, the distributions for both community categories nearly reached their 100\%.
Overall, we found that AI-claiming comments tend to emerge rapidly, and particularly in \textit{AI-disallowed} communities, where they appear even more swiftly.




The time dynamics show a different response to suspected images in the two types of communities. 
\newt{Furthermore, t}o better understand the response, we conducted a sentiment analysis of comments suggesting AI involvement in the \textit{Suspected AI posts}.   
To label the sentiment of comments implying AI, we used a large language model as a classifier, following a similar approach to that described in the previous sections, which is based on prior studies~\cite{wei2022chain,rao2024quallm}.
We used the LLM classifier to label the comments as  Negative, Neutral, or Positive, using a prompt that can be found in Appendix~\ref{llm_prompts_sa}, along with examples of input texts and their corresponding classifications.
To assess the accuracy of this method, we manually labeled a random sample of 100 comments and achieved an accuracy rate of 92\% for this sentiment classification task.

Using these labels, we looked at differences in comment sentiment between \textit{AI-disallow} and \textit{AI-neutral} communities, and how they may have changed over time.
To do that, we compared sentiment in two distinct time periods: January-March 2023 and October-December 2023 (the end of our data).
In January-March 2023, the percentage of negative sentiment in comments in \textit{AI-disallowing communities} was 39\% (6\% were positive, and the rest neutral).
The ratio of negative sentiment was lower in \textit{AI-neutral} communities, at around 28\%.
By October-December 2023, however, the pattern reversed, with \textit{AI-neutral} communities exhibiting a higher percentage of negative comments (51\%) than AI-disallowing communities (42\%).
The reason for the flip could be the better control exercised by the \textit{AI-disallowing communities} about AI posts\newt{, or a normative shift in the \textit{AI-neutral communities} with respect to AIGC}.
Still, the data shows an increase over time in the proportion of comments expressing negative sentiment among those claiming AI involvement, across both community categories.


\section{Discussion}
\label{sec:disc}
Our study provides insights about how AI-generated content is impacting online communities, specifically those that deal with creative visual content -- in this case, art-related subreddits. 

Perhaps surprisingly, we observed low volumes of (admitted, transparent) AI-generated content in these communities compared to other media-based posts (RQ1). 
Posts that either transparently used AI or were accused of doing so, together amounted to less than 0.5\% of media posts.
Of course, it is likely that AI-generated image posts are more prevalent, but neither the authors nor the commenters are identifying them.
Studies on the prevalence of AIGC online are sparse, but those that exist in other domains for now also indicate low numbers -- for example, estimating 1.39\% of machine-generated news articles for mainstream news websites~\cite{hanley2024machine} or up to 0.05\% of AI-generated profiles on Twitter~\cite{ricker2024ai}.
Of course, since this technology is still new, any presence of AIGC on public platforms points towards propagation in the future, and a fundamental shift to how people communicate digitally.

Interestingly, though, our data reveals that the volume of \textit{Transparent AI posts} (RQ1) decreased over time after a first (and expected) rise, perhaps due to declining novelty, changes in community policies and values, or shifts in how users present the content and respond to it.
Most significantly, this decrease could be due to posters being more reluctant to admit their posts are AI-generated, as community rules and community and social norms around AI use are changing~\cite{lloyd2025airules}.  
Indeed, responses to posts that are suspected of using AI rose later, and remained somewhat higher through the end of our data collection.
These dynamics suggest that we have entered a new era in online communities, where the public is suspicious about the AI provenance of information.
Unfortunately, we know that suspicion of AI content is likely to lower trust online more broadly~\cite{jakesch2023human}.
We also find that people may be becoming more negative about AIGC over time, as negative sentiments in comments for \textit{Suspected AI posts} increased between early and late 2023. 

In general, these findings point to the dynamic nature of rules and norms in online communities.
There is clearly a difference in how communities decided to tackle the challenge of AI-generated content, and research shows that these norms and rules about AI have both changed over time~\cite{lloyd2025airules}. 
It is important to continue to understand what brings about change of rules, how they are implemented, and how norms might shift as a result.
Prior work identified that rules forbidding certain actions may appear later in a subreddit's lifecycle and are often associated with larger subreddits~\cite{reddy2023evolution, fiesler2018reddit}, and that this trend is similar for rules about AI-generated content~\cite{lloyd2025airules}.
While we used a binary classification system to differentiate between \textit{AI-disallowed} and \textit{AI-neutral} communities, it is possible and even likely that our \textit{AI-disallowed} communities implemented rules forbidding AI at some point in the middle of our dataset timespan.
This possibility may underrepresent the strength of some of our comparative findings where, for example, \textit{AI-disallowed communities} have not actually disallowed AI for the full period.
There is a rich space for future work in this area of shifting rules and norms.
For example, scholars could study communities immediately before and after a rule change forbidding AI is implemented, and investigate the impact on content posting.
Future work may also look to evaluate the effects of \textit{how} subreddit bans are implemented and enforced.
For example, prior work in other settings has found that exposing rulesets more readily can decrease harmful content while increasing newcomer participation~\cite{matias2019preventing}.

Our findings carry implications for how AI may enable newcomers to participate in online communities, but also how AI-interested newcomers may change existing communities (RQ2).
First, we find that newcomers to communities create proportionally more \textit{Transparent AI posts} than general image-based posts.
This newcomer over-representation confirms prior work that suggests that the introduction of generative AI leads to more participation and posting~\cite{guo2024exploring, wei2024understanding, zhou24art}.
We show that users whose first image-based post to a subreddit is a self-labeled AI-generated image continue to post such AIGC at high rates within the subreddit.
This behavior suggests that creators who initially joined these subreddits to transparently share AI-generated art, and have returned to post again, are truly interested in AI-generated images---or find it as their main possible avenue to contribute.
Kraut et al.~\cite{kraut2012challenges} identify five basic challenges that online communities face when dealing with newcomers, and our findings suggest that while the ready availability of AIGC may help with the \textit{recruitment} and \textit{retention} of new members, these new members do not become \textit{socialized} in the exact same way as members that do not post AIGC.
Instead, new members that post AIGC appear to wish to continue posting AIGC and may become turned off from online communities once AI content is banned.
Indeed, we find evidence that when such newcomers post to subreddits where AI-generated content is prohibited, they are less likely to remain engaged within the community.
Possible explanations for this lower retention include criticism or bans from the moderators of the subreddits, negative feedback from other participants, low interaction or engagement with their posts, or the possibility that these users are low-value contributors that are less invested in the community itself and thus contribute less frequently.
Of course, lower retention of newcomers may not be a universal negative if new contributions are low-effort, and prior work has found moderators are sometimes willing to make these sacrifices to uphold community values~\cite{lloyd2023there}.
Our analysis of the response to content (RQ3) suggests that high levels of community engagement on the post (e.g. comments) are associated with AI suspicion by commenters, in what we may want to call \textit{Botwin's Law} (with apologies to Mr. Godwin): the more discussion or exposure an item gets, the more likely it is to be suspected of using AI. 
Or, in other words, AI suspicion may be more readily attached to successful posts. 
Our findings also suggest that AI accusations arrive rapidly, especially in \textit{AI-disallowed communities}, reflecting the heightened awareness among participants regarding the possibility of generative art.
This rapid response could indicate the role of moderators or community members working to uphold the standards and environment of their respective communities by quickly identifying and challenging suspected AI-generated works.

We note that this work comes with some potential limitations.
Of course, as we only looked at art communities, our findings mostly apply to this specific type of community and may not be broadly applicable.
For example, perceptions, responses, and use of AIGC are likely to be different in non-visual or non-creative settings.
Additionally, while we looked at 57 different subreddits, it is possible that our findings were disproportionately influenced by the most dominant communities---although we did our best to verify that this imbalance did not sway the results. 
For example, we verified that the observed trends in the timelines were not disproportionately influenced by the larger subreddits.
However, potential biases toward more prominent or active communities could affect our conclusions, particularly if these dominant subreddits exhibit unique community rules, cultural norms, or dynamics.
Additionally, only data that was included in the PushShift torrents is included in this dataset, which overrepresents highly popular subreddits.
Finally, the constraints of dataset collection mean that it is possible that posts may have been deleted before they could be gathered by the torrents, though evidence suggests that PushShift tends to retrieve data quickly~\cite{reagle2023even}.

The data processing and analysis steps may also have certain limitations, which could be addressed in future research.
Identifying posts containing AI-generated media relies on text-based methods, which sometimes may capture user opinion and reactions, rather than accurately reflect the actual prevalence of AI-generated content.
For example, in \textit{AI-neutral} subreddits where AI-generated content is permitted, users can be less inclined to explicitly mention it in posts or comments.
In contrast, in \textit{AI-disallowed} subreddits, users may be more likely to claim that others' content is AI-generated as it violates community rules.
In the categorization of the communities step, given that \textit{AI-neutral communities} do not explicitly address AI usage in their guidelines or descriptions, while \textit{AI-disallowed communities} explicitly prohibit it, the observed differences in trends between these groups may reflect broader factors, such as more adaptive approach to emerging technologies by the \textit{AI-disallowed communities}. 
That is, factors beyond the subreddit-specific policies regarding AI-generated content.
Additionally, the analysis may face potential limitations related to the identification of newcomers, as the dataset may include bot or troll accounts considered as newcomers. Bots, in particular, are more likely to use AI-generated content~\cite{grimme2022new,ferrara2023social}, and may be at higher risk of suspension by the platform. During data collection, measures were taken to filter out potential bot accounts based on usernames and repeated comments. 
However, these efforts may not have entirely excluded all accounts exhibiting bot-like behavior.
\newt{Finally, future work could refine the sentiment analysis by moving beyond the three-label scheme (Negative, Neutral, Positive) to provide more nuanced insights.}

In summary, our findings correspond to a period of rapid advancements in AI-generated art technologies, which increasingly are able to produce high-quality content resembling traditional art forms.
With this change, moderators of online communities contend with old challenges, such as whether appealing to newcomers is worth compromising on content quality, but are also faced with decisions that may fundamentally compromise their community values.
Regardless of which rule sets communities choose to adopt, our findings show that suspicion and negativity over AI-generated content is here to stay.
As the prevalence of AI-generated content grows, we must find ways to preserve and center core community values -- such as creativity, authenticity, and openness -- in social online ecosystems.

\bibliographystyle{ACM-Reference-Format}
\bibliography{bib}

\appendix

\section{List of Subreddits}
\label{list_subreddits}

\begin{itemize}
\item AbstractArt
\item alternativeart
\item AmateurArt
\item AnimeART
\item AnimeSketch
\item Art
\item ArtBuddy
\item ArtCrit
\item ArtDeco
\item ArtHistory
\item ArtPorn
\item ArtProgressPics
\item ArtSale
\item artstore
\item BadArt
\item Beginner\_Art
\item comicbookart
\item comic\_crits
\item conceptart
\item ContemporaryArt
\item CreepyArt
\item CryptoArt
\item DarkGothicArt
\item DeviantArt
\item DigitalArt
\item DrawForMe
\item drawing
\item DrugArt
\item FantasyArt
\item FractalPorn
\item FurryArtSchool
\item generative
\item glitchart
\item Graffiti
\item HungryArtists
\item IDAP
\item Illustration
\item ImaginaryCharacters
\item ImaginaryLandscapes
\item ImaginaryMindscapes
\item ImaginaryMonsters
\item ImaginaryTechnology
\item learnart
\item painting
\item photoshopbattles
\item picrequests
\item PixelArt
\item redditgetsdrawn
\item Sculpture
\item sketchpad
\item SpecArt
\item stencils
\item streetart
\item TransformersArt
\item UnusualArt
\item Watercolor
\item wtfart
\end{itemize}

\section{List of Keywords}
\label{list_keywords}
\begin{itemize}
\item {['ai-generated']}
\item {['generated', 'ai']}
\item {['generated', 'artificial', 'intelligence']}
\item {['created', 'artificial', 'intelligence']}
\item {['created', 'ai']}
\item {['ai-created']}
\item {['ai-powered']}
\item {['dall-e']}
\item {['dalle']}
\item {['midjourney']}
\item {['stable', 'diffusion']}
\item {['made', 'ai']}
\item {['neural', 'network', 'generated']}
\item {['gan', 'generated']}
\item {['deep', 'learning', 'generated']}
\item {['dl', 'generated']}
\item {['machine', 'learning', 'generated']}
\item {['ml', 'generated']}
\item {['artbreeder']}
\item {['deepdream']}
\item {['ai', 'art']}
\item {['ai', 'painting']}
\item {['wombo']}
\item {['starryai']}
\item {['nightcafe']}
\item {['deepai']}
\item {['craiyon']}
\item {['vqgan+clip']}
\item {['ai', 'generative']}
\item {['ai', 'generation']}
\item {['generating', 'ai']}
\item {['generativeai']}
\item {['ai', 'image', 'generator']}
\item {['artificial', 'intelligence', 'generative']}
\item {['artificial', 'intelligence', 'generation']}
\item {['made', 'artificial', 'intelligence']}
\item {['neural', 'network', 'generation']}
\item {['deep', 'learning', 'generation']}
\item {['dl', 'generation']}
\item {['ai', 'image']}
\item {['artificial', 'intelligence', 'image']}
\item {['ai', 'artwork']}
\item {['artificial', 'intelligence', 'artwork']}
\item {['artificial', 'intelligence', 'art']}
\item {['aiart']}
\item {['gan', 'art']}
\item {['neural', 'network', 'art']}
\item {['painted', 'ai']}
\item {['painted', 'artificial', 'intelligence']}
\item {['artificial', 'intelligence', 'painting']}
\item {['artificial', 'intelligence', 'paint']}
\item {['artificial', 'intelligence', 'drawing']}
\item {['artificial', 'intelligence', 'draw']}
\item {['ai', 'paint']}
\item {['ai', 'drawing']}
\item {['ai', 'draw']}
\item {['drew', 'ai']}
\item {['drew', 'artificial', 'intelligence']}
\item {['deepdreamgenerator']}
\item {['stablediffusion']}
\item {['stable\_diffusion']}
\item {['dall', 'e']}
\item {['dall-e2']}
\item {['dall-e3']}
\item {['wombodream']}
\item {['invokeai']}
\item {['dcgan']}
\item {['stylegan']}
\end{itemize}

\section{LLM Prompts for Data Classification}
\label{llm_prompts_classification}

\subsection{Title or Body of the Post}
Task:
\newline Process a list of texts from art-related posts on Reddit, where each post includes an attached image or video. The provided text is either the title of the post, or its body.
\newline Your task is to classify the texts based on whether they suggest the attached images are AI-generated.
\\
\\
The classification categories are:
\newline A. Imply AI involvement: The text explicitly or implicitly suggests that AI tools (such as machine learning, deep learning, neural networks, GANs, or text-to-image solutions) were involved at some stage in generating the attached image.
\newline B. Does not imply AI: The text mentions AI-generated images or tools for generating AI art, but does not imply that the attached image is AI-generated. This includes texts that state that the attached image is not AI-generated, or texts that explain that the image was created using traditional art methods (without AI), or texts that discuss AI tools or AI-generated art in general without associating them with the attached image.
\newline C. Unrelated to AI: Choose this option only if the text is completely unrelated to AI-generated images, and completely unrelated to tools for generating AI art. Apply this category if the text does not mention AI-generated images or tools at all, regardless of whether the text refers to the attached image.
\newline Note: If the text does not refer to the attached image at all, then you must choose either B or C (not A). Then, only if the text does not mention AI-generated images or related tools in any context, then you must choose C (not B).
\\
\\
Input format:
\newline A JSON-style dictionary with a serial index as the key, and the text as the value.
\\
\\
Process:
\newline 1. For each text, concisely provide your reasoning, describe how you analyzed the content, and explain your thought process leading to the final classification. Output your reasoning only, without repeating the input text itself.
\newline 2. After your reasoning, provide the final classification for this text. Output category identifier only (A, B, or C), without repeating its name or explanation.
\newline Note: Remember to provide reasoning for each text before giving the final classification. Analyze each text thoroughly, considering multiple aspects before making a decision. 
\\
\\
Output format:
\newline Combine the reasoning and classification into a single output for each text. Present the final output as a JSON-style dictionary with the serial index as the key, and a list as the value. This list should contain your reasoning as the first element, and the final classification as the second element.
\newline Example of the final output format:
\newline \{1: [``Reasoning for text 1'', ``A''],
\newline  2: [``Reasoning for text 2'', ``A''],
\newline  3: [``Reasoning for text 3'', ``B'']\}
\newline Note: Ensure that every text in the input list has a corresponding classification. Each text should be classified under one category only.
\newline Pay attention to the structure of the dictionary: Place a `:' between each key and its value, use a list of length 2 as a value, use a `,' to separate each key-value pair (but do not place a comma after the last pair), begin the dictionary with `\{' and end it with `\}'.
\newline Do not generate any other text in the final output dictionary. 
\\
\\
In-context examples:
\newline Example input:
\newline \{1: Starting to train an Artificial Neural Network to generate images related to the Google search term ``Ayahuasca Art'',
\newline 2: Genesis In Blue This as an Abstract painting of mine about the Birth of Artificial Intelligence... Copies are available for sale on Fine Art America...,
\newline 3: ``The skin of the city.'' Machine learning generated city textures. (abstract art),
\newline 4: Life and Death, Midjourney, Digital, 2022,
\newline 5: Magical princess, I draw on an old discarded sketch in studio, by me, no AI,
\newline 6: Haters will say it's AI art,
\newline 7: Soldier Broadcast, The Broadcaster, Pencil drawing enhanced with AI, 2021,
\newline 8: Can somebody make the hands on these look normal? (Yes these are Ai generated, hope that’s not an issue),
\newline 9: Street art in Ramat Gan, Israel\}
\\
\\
Example output:
\newline \{1: [``The text explicitly mentions starting to train an artificial neural network to generate images, indicating the use of AI tools.'', ``A''],
\newline  2: [``The text describes an abstract painting, without mentioning AI tools used to create it. The term ``Artificial Intelligence'' in the title seems to refer to the theme rather than the creation process.'', ``C''],
\newline  3: [``The text clearly states that machine learning was used to generate city textures, indicating AI involvement.'', ``A''],
\newline  4: [``The text mentions ``Midjourney'' as part of the characteristics that describe the image. Midjourney is a known AI tool for generating images, indicating that the attached image is AI-generated.'', ``A''],
\newline  5: [``The text specifies that the drawing was done by the person and explicitly states ``no AI'', indicating that AI was not used.'', ``B''],
\newline  6: [``The text suggests that people might wrongly assume the artwork is AI-generated, implying it is not AI-generated.'', ``B''],
\newline  7: [``The text indicates that a pencil drawing was enhanced with AI, clearly indicating AI involvement.'', ``A''],
\newline  8: [``The text explicitly states that the images are AI generated.'', ``A''],
\newline  9: [``The text simply describes street art in a specific location without mentioning AI or any claims regarding the generation method.'', ``C'']\}
\\
\\
Input:
\newline \{\textit{input\_dictionary}\}
\\
\\
Provide your output as specified.

\subsection{Comments Made by the Author of the Post}
Task:
\newline Process a list of texts from art-related posts on Reddit, where each post includes an attached image or video. The provided text is a comment made by the author of the post. 
\newline Your task is to classify the comments based on whether they claim the attached images are AI-generated.
\\
\\
The classification categories are:
\newline A. Imply AI involvement: The comment explicitly or implicitly suggests that AI tools (such as machine learning, deep learning, neural networks, GANs, or text-to-image solutions) were involved at some stage in generating the image in the post. Apply this category if the author implies that AI was used to create the image, or if the author implies that the image is an output of AI tools.
\newline B. Does not imply AI: The comment mentions AI-generated images or tools for generating AI art, but does not imply that the image in the post is AI-generated. This includes comments that state that the image in the post is not AI-generated, or comments that explain that the image was created using traditional art methods (without AI), or comments that discuss AI tools or AI-generated art in general without associating them with the image in the post. You can choose this option regardless of whether the text refers to the image in the post, or not.
\newline C. Unrelated to AI: Choose this option only if the text is completely unrelated to AI-generated images, and completely unrelated to tools for generating AI art. Apply this category if the text does not mention AI-generated images or tools at all, regardless of whether the text refers to the image in the post.
\newline Note: If the comment does not refer to the image in the post at all, then you must choose either B or C (not A). Then, only if the comment does not mention AI-generated images or related tools in any context, then you must choose C (not B).
\\
\\
Input format:
\newline A JSON-style dictionary with a serial index as the key, and the comment as the value. 
\\
\\
Process:
\newline 1. For each comment, concisely provide your reasoning, describe how you analyzed the content, and explain your thought process leading to the final classification. Output your reasoning only, without repeating the input text itself.
\newline 2. After your reasoning, provide the final classification for this comment. Output category identifier only (A, B, or C), without repeating its name or explanation.
\newline Note: Remember to provide reasoning for each comment before giving the final classification. Analyze each comment thoroughly, considering multiple aspects before making a decision.
\\
\\
Output format:
\newline Combine the reasoning and classification into a single output for each comment. Present the final output as a JSON-style dictionary with the serial index as the key, and a list as the value. This list should contain your reasoning as the first element, and the final classification as the second element.
\newline Example of the final output format:
\newline \{1: [``Reasoning for comment 1'', ``A''],
\newline  2: [``Reasoning for comment 2'', ``A''],
\newline  3: [``Reasoning for comment 3'', ``B'']\}
\newline Note: Ensure that every comment in the input list has a corresponding classification. Each comment should be classified under one category only.
\newline Pay attention to the structure of the dictionary: Place a `:' between each key and its value, use a list of length 2 as a value, use a `,' to separate each key-value pair (but do not place a comma after the last pair), begin the dictionary with `\{' and end it with `\}'.
\newline Do not generate any other text in the final output dictionary. 
\\
\\
In-context examples:
\newline Example input:
\newline \{1: Been attempting to create organic looking images with various electronic tools. The source images were a touchdesigner patch being routed through an analog video mixer. I then took those images and ran them through some AI which created some interesting image sequence. That then became this - through various morphing, blurring and stretching.,
\newline 2: This was made with an AI and a text prompt. I used VQGAN+CLIP for it,
\newline 3: NO AI is ever used in my work. Created by hand in procreate on an iPad. Took about 7 hours to create. This one was inspired by my frustration when others try to dictate the value of an artist. We need to find our own value and self-worth. I am a huge advocate for artist support especially in today's climate as the art world is changing so much.,
\newline 4: This is not AI generated. I drew it using Rebelle 3 but I do not have a timelapse,
\newline 5: It is , VQGAN is AI making art from words you choose .,
\newline 6: I know you guys may think this is AI but it’s totally my original work, it took me hours to paint this. Don’t hate because you lack talent,
\newline 7: Thank you for checking this piece out! Here’s my insta if you’d like to see more: https://www.instagram.com/ai.shii.art/,
\newline 8: First made dinosaur in AI, then i used PS to make it look like pixel art, its so much easier then doing ``real'' pixel art.,
\newline 9: https://www.deviantart.com/print/view/887902254
\newline https://www.pinterest.de/pin/1136666393422007051
\newline https://www.artstation.com/prints/art\_poster/j0oV/21-690-deepdream-art\}
\\
\\
Example output:
\newline \{1: [``Comment indicates the use of AI in creating the image.'', ``A''],
\newline  2: [``Comment explicitly mentions the use of AI and specific AI tools to create the image.'', ``A''],
\newline  3: [``Comment explicitly states no AI was used and describes traditional art creation methods.'', ``B''],
\newline  4: [``Comment explicitly states the image is not AI-generated.'', ``B''],
\newline  5: [``Comment indicates the use of AI for generating the image from words.'', ``A''],
\newline  6: [``Comment explicitly states that the image is not AI-generated, but an original work.'', ``B''],
\newline  7: [``Comment is unrelated to AI-generated images or to tools for generating AI art.'', ``C''],
\newline  8: [``Comment indicates that AI was used at the initial stage of creating the image.'', ``A''],
\newline  9: [``This comment contains links related to AI art resources and guidance, without mentioning AI-generated images or methods to generate images.'', ``C'']\}
\\
\\
Input:
\newline \{\textit{input\_dictionary}\}
\\
\\
Provide your output as specified.

\subsection{Comments Made by Other Commenters}
Task:
\newline Process a list of texts from art-related posts on Reddit, where each post includes an attached image or video. The provided text is one of the comments on the post. 
\newline Your task is to classify the comments based on whether they suggest the attached images are AI-generated.
\\
\\
The classification categories are:
\newline A. Imply AI involvement: The comment explicitly or implicitly suggests that AI tools (such as machine learning, deep learning, neural networks, GANs, or text-to-image solutions) were involved at some stage in generating the image in the post. Apply this category if the commenter ultimately believes or suspects that AI was used to create the image, or if the commenter claims that the image looks as the output of AI tools.
\newline B. Does not imply AI: The comment mentions AI-generated images or tools for generating AI art, but does not imply that the image in the post is AI-generated. This includes commenters that state that the image in the post is not AI-generated, or commenters that only initially thought the image might be AI-generated, or commenters that discuss AI tools or AI-generated art in general without associating them with the image in the post. This also includes comment that explain that the submission breaks the rules of the subreddit, where posting AI images is one of the possible accusations. 
\newline C. Unrelated to AI: Choose this option only if the text is completely unrelated to AI-generated images, and completely unrelated to tools for generating AI art. Apply this category if the text does not mention AI-generated images or tools at all, regardless of whether the text refers to the image in the post.
\newline Note: If the comment does not refer to the image in the post at all, then you must choose either B or C (not A). Then, only if the comment does not mention AI-generated images or related tools in any context, then you must choose C (not B).
\\
\\
Input format:
\newline A JSON-style dictionary with a serial index as the key, and the comment as the value. 
\\
\\
Process:
\newline 1. For each comment, concisely provide your reasoning, describe how you analyzed the content, and explain your thought process leading to the final classification. Output your reasoning only, without repeating the input text itself.
\newline 2. After your reasoning, provide the final classification for this comment. Output category identifier only (A, B, or C), without repeating its name or explanation.
\newline Note: Remember to provide reasoning for each comment before giving the final classification. Analyze each comment thoroughly, considering multiple aspects before making a decision.
\\
\\
Output format:
\newline Combine the reasoning and classification into a single output for each comment. Present the final output as a JSON-style dictionary with the serial index as the key, and a list as the value. This list should contain your reasoning as the first element, and the final classification as the second element.
\newline Example of the final output format:
\newline \{1: [``Reasoning for comment 1'', ``A''],
\newline  2: [``Reasoning for comment 2'', ``A''],
\newline  3: [``Reasoning for comment 3'', ``B'']\}
\newline Note: Ensure that every comment in the input list has a corresponding classification. Each comment should be classified under one category only.
\newline Pay attention to the structure of the dictionary: Place a `:' between each key and its value, use a list of length 2 as a value, use a `,' to separate each key-value pair (but do not place a comma after the last pair), begin the dictionary with `\{' and end it with `\}'.
\newline Do not generate any other text in the final output dictionary. 
\\
\\
In-context examples:
\newline Example input:
\newline \{1: First I thought that its some heavily-processed image (like deepdream). But then I read the title and I'm like :O Great work!,
\newline 2: Heyy, I know how you did that. You used an app called Wombo Dream that uses GAN AI to generate art from a prompt. I love that app. I'm guessing the style you used is `psychic'.,
\newline 3: I'm calling bullshit. If you look at OP's posts, their other art is EXTREMELY different than this. Also, if you zoom in, there's NO WAY OP did this with colored pencils...there are individual hairs that are just too perfect. I'm guessing OP used a colored pencil filter over a photo or used AI to generate it.,
\newline 4: Hey there I do a bunch of atuff like this if you like my art ai would gladly make this piece for you www.instagram.com/mental\_foundry/,
\newline 5: Omg this looks like something from Artbreeder!!!,
\newline 6: Using AI obviously lets you stay competitive in terms of image reproduction :| but not creative. If it’s what you love doing, go for it  ;),
\newline 7: Is this AI generated?,
\newline 8: This wasn’t made with AI, the characters name is Ai,
\newline 9: It's also AI art which isn't allowed, this is the fourth one the OP has posted in the last two days and the previous three were already removed for being AI generated as well.\}
\\
\\
Example output:
\newline \{1: [``The commenter initially thought the image might be AI-generated (deepdream) but later concluded it isn't AI-generated after reading the title. This implies an initial suspicion but concludes that AI was not used.'', ``B''],
\newline 2: [``The commenter explicitly states the image is generated using an AI application called Wombo Dream.'', ``A''],
\newline 3: [``The commenter expresses strong skepticism about the image being authentically drawn and suggests AI involvement.'', ``A''],
\newline 4: [``This comment includes an ambiguous reference to ``ai'' which could be read as `I' or `AI', but given the context, it likely refers to "I" (the artist themselves) and does not mention AI involvement.'', ``C''],
\newline 5: [``The commenter states the image looks like it could be generated from Artbreeder, a known AI tool. This suggests a suspicion that AI might have been used for the image.'', ``A''],
\newline 6: [``The commenter mentions AI generation in a general context, discussing competitiveness and creativity but does not imply the post's image was AI-generated.'', ``B''],
\newline 7: [``The commenter is directly asking if the image is AI-generated, indicating a suspicion of AI involvement.'', ``A''],
\newline 8: [``The commenter clarifies that AI was not used, explaining that the character's name is ``Ai.'''', ``B''],
\newline 9: [``The commenter explicitly states that the image is AI-generated.", ``A'']\}
\\
\\
Input:
\newline \{\textit{input\_dictionary}\}
\\
\\
Provide your output as specified.

\section{LLM Prompt for Sentiment Analysis}
\label{llm_prompts_sa}

Task:
\newline Process a list of texts from art-related posts on Reddit, where each post includes an attached image or video.
\newline The provided text is one of the comments on the post. This comment implies that the image in the post was AI-generated.
\newline Your task is sentiment analysis - classify the provided comments based on their sentiment.
\\
\\
The classification categories are:
\newline A. Negative - The commenter expresses a negative sentiment in their response. For example, they might be disappointed, blaming, angry, or dismissive. This includes commenters who belittling the creator of the image or the attached image itself, or claiming they do not consider the attached image as art, or imply that it is not authentic art.
\newline B. Neutral - The commenter simply asks, implies, or claims that the attached image was created by AI, without expressing a negative or positive sentiment.
\newline C. Positive - The commenter reacts positively to the attached image, for example, by complimenting the creator, or expressing that they are impressed by the image.
\\
\\
Input format:
\newline A JSON-style dictionary with a serial index as the key, and the comment as the value. 
\\
\\
Process:
\newline 1. For each comment, concisely provide your reasoning, describe how you analyzed the content, and explain your thought process leading to the final classification. Output your reasoning only, without repeating the input text itself.
\newline 2. After your reasoning, provide the final classification for this comment. Output category identifier only (A, B, or C), without repeating its name or explanation.
\newline Note: Remember to provide reasoning for each comment before giving the final classification. Analyze each comment thoroughly, considering multiple aspects before making a decision.
\\
\\
Output format:
\newline Combine the reasoning and classification into a single output for each comment. Present the final output as a JSON-style dictionary with the serial index as the key, and a list as the value. This list should contain your reasoning as the first element, and the final classification as the second element.
\newline Example of the final output format:
\newline \{1: [``Reasoning for comment 1'', ``A''],
\newline  2: [``Reasoning for comment 2'', ``B''],
\newline  3: [``Reasoning for comment 3'', ``B'']\}
\newline Note: Ensure that every comment in the input list has a corresponding classification. Each comment should be classified under one category only.
\newline Pay attention to the structure of the dictionary: Place a `:' between each key and its value, use a list of length 2 as a value, use a `,' to separate each key-value pair (but do not place a comma after the last pair), begin the dictionary with `\{' and end it with `\}'.
\newline Do not generate any other text in the final output dictionary. 
\\
\\
In-context examples:
\newline Example input:
\newline \{1: This has a lot of AI generated vibes, love it!,
\newline 2: i feel like it’s AI generated,
\newline 3: Its AI generated, so congratulate the machine.,
\newline 4: You did not make that. AI made it.,
\newline 5: Not really glitch, just a deepdream.,
\newline 6: Omg this looks like something from Artbreeder!!!,
\newline 7: Looks like its either stolen or AI "art",
\newline 8: This looks very AI-ish, Midjourney?,
\newline 9: AI generated content is not allowed on this subreddit\}
\\
\\
Example output:
\newline \{1: [``The commenter acknowledges the AI-generated vibes but expresses love for the image, indicating a positive sentiment.'', ``C''],
\newline  2: [``The commenter neutrally suggests that the image feels AI-generated without further sentiment.'', ``B''],
\newline  3: [``The commenter states that the image is AI-generated, and sarcastically congratulates the machine used to generate it.'', ``A''],
\newline  4: [``The commenter directly accuses the poster of not creating the image and implies it was made by AI, expressing a dismissive and negative sentiment.'', ``A''],
\newline  5: [``The commenter offers a neutral observation that the image is not really glitch art but is similar to deepdream, without expressing any strong sentiment.'', ``B''],
\newline  6: [``The commenter expresses excitement by exclaiming that the image resembles something from Artbreeder, which suggests a positive sentiment.'', ``C''],
\newline  7: [``The commenter suggests the image looks stolen or AI-generated, implying suspicion and a negative view of the image.'', ``A''],
\newline  8: [``The commenter observes that the image looks AI-generated, mentioning Midjourney, without expressing further sentiment.'', ``B''],
\newline  9: [``The commenter states that AI-generated content is not allowed on the subreddit, which implies a negative sentiment towards AI-generated content.'', ``A'']\}
\\
\\
Input:
\newline \{\textit{input\_dictionary}\}
\\
\\
Provide your output as specified.

\end{document}